\documentclass[prb,10pt, print,twocolumn]{revtex4-1}

\usepackage[utf8]{inputenc}
\usepackage{xcolor}
\usepackage{comment}
\usepackage{xcolor,soul,ulem}
\usepackage{amsmath}
\usepackage{amssymb}
\usepackage{array}
\usepackage{graphicx}

\usepackage[caption=false]{subfig}

\newenvironment{conditions}[1][Where:]
  {#1 \begin{tabular}[t]{>{$}l<{$} @{${}={}$} l}}
  {\end{tabular}\\[\belowdisplayskip]}
\newcommand{\R}{\mathbb{R}}

\begin{document}

\title{Modular Multimodal Architecture for Document Classification.}
\author{T. Dauphinee, N. Patel, M. Rashidi}

\begin{abstract}
    Page classification is a crucial component to any document analysis system, allowing for complex branching control flows for different components of a given document. Utilizing both the visual and textual content of a page, the proposed method exceeds the current state-of-the-art performance on the RVL-CDIP benchmark at 93.03\% test accuracy.
\end{abstract}
\maketitle

\section{Introduction}
The promise of a paperless society much like cold fusion and flying cars, has always been  a decade away, and although we do come closer to this dream by the day, there are many industries that still rely heavily on paper based processes. Human communication is vague and context based, even given a precisely structured form to fill out, people find a way to subvert the structure to convey what they really want to say. This tendency to color outside the lines makes automating paper based processes difficult, however recent advances in computer vision and natural language processing have pushed us further toward that utopian paperless future.

Document analysis systems follow a general abstract structure, with three overall components: text/structure extraction, page classification/sorting, and content understanding. In a commercial setting there is almost always extraneous and possibly confounding information contained within a submitted document, requiring a robust and flexible way to classify the pages of a given document to ensure the isolation of the correct source of information. This is especially relevant when building a larger pipeline where downstream processes rely on the results of page classification. In these situations, a small incremental boost in classification performance can net much larger performance boosts for the overall pipeline.

The topic of document page image classification has received much publicity over the last few years. In fact the RVL-CDIP\cite{harley2015icdar} dataset was curated specifically to test image classification strategies on document images. Earlier studies focused heavily on the original AlexNet \cite{Krizhevsky:2012:ICD:2999134.2999257} architecture \cite{harley2015icdar,tensmeyer2017analysis}. More recently modern architectures such as VGG16 \cite{simonyan2014deep}, GoogLeNet \cite{DBLP:journals/corr/SzegedyLJSRAEVR14}, and ResNet50\cite{DBLP:journals/corr/HeZRS15} have been proposed and tested on RVL-CDIP \cite{Afzal_2017}. The current state-of-the-art utilizes a set of 5 distinct VGG16 models, one for the whole image (known as the holistic model, initialized with pretrained ImageNet weights) and 4 for specific subsections of the image (header, footer, left body, and right body initialized on the holistic trained weights). These 5 models are then combined to form a final prediction \cite{das2018document}. While accurate, the number of parameters is immense (on the order of $10^{8}$) and the training process is sequential, requiring a holistic model to be trained before any of the subsection models can be trained.

In addition to the aforementioned image classification strategies, we can take advantage of optical character recognition (OCR) technology to extract text from document page images and train text classification algorithms. Modern OCR systems are not infallible, especially in the context of low quality scanned documents. Typically the output of an OCR system will contain transcription errors (ex. mistaking i for l and vice versa) due to noise in the source image. Many approaches have been developed to deal with text classification problems \cite{DBLP:journals/corr/abs-1904-08067}, although most have been developed under the assumption of clean encoded text. There is evidence to support that the bag-of-words approach is quite robust to the unavoidable transcription errors \cite{vinciarelli2005noisy, agarwal2007much}. To the best of our knowledge, there are no studies showing similar analysis for word embedding methods, however it can by hypothesized that transcription errors are amplified in the embedding space, opening an avenue for future research.

Given both the image and text classification approaches, it is natural to design a system that combines both to form a joint modelling approach, typically referred to as a multimodal classification model. This is not a new idea and in fact we find literature dating back before the explosion in popularity of convolutional neural networks (CNN) for image classification. \cite{augereau2014improving}. More recently a study has been conducted utilizing a similar procedure to the proposed work, with a focus on minimal model footprint in commercial application \cite{audebert2019multimodal}. Another commercial study was also conducted, utilizing the proposed abstract structure with a private dataset \cite{enginmultimodal}. Both of these studies suggest that adding text information improves model performance substantially.

 In this work we explore combining both approaches into a single classification task, i.\,e. we construct a model that uses both the visual information and the textual content of page to make a decision. To test the proposed architecture we take advantage of an open and freely available dataset, RVL-CDIP \footnote{https://www.cs.cmu.edu/~aharley/rvl-cdip/}. We show that the proposed method exceeds the current state-of-the-art performance on this dataset with a test accuracy of 93.03\%.

\section{Proposed Method}

\subsection{Text Extraction}
We utilize the open source \footnote{https://github.com/tesseract-ocr/tesseract} Tesseract OCR engine \cite{smith2007overview} to extract text from all images in the RVL-CDIP dataset. It is important to note that the only preprocessing step involved is resizing such that the longest dimension is 3300 pixels. This choice was made to ensure conformity to the suggested minimum DPI of 300 with the assumption that every page is standard letter size (this appears to generally be true for this dataset). We use the the combined legacy/LSTM engine (\texttt{oem 3}) and the standard page segmentation mode (\texttt{psm 3}) parameters for this extraction. 

\subsection{Abstract Model Architecture}
\begin{figure}
    \includegraphics[width=7.5cm]{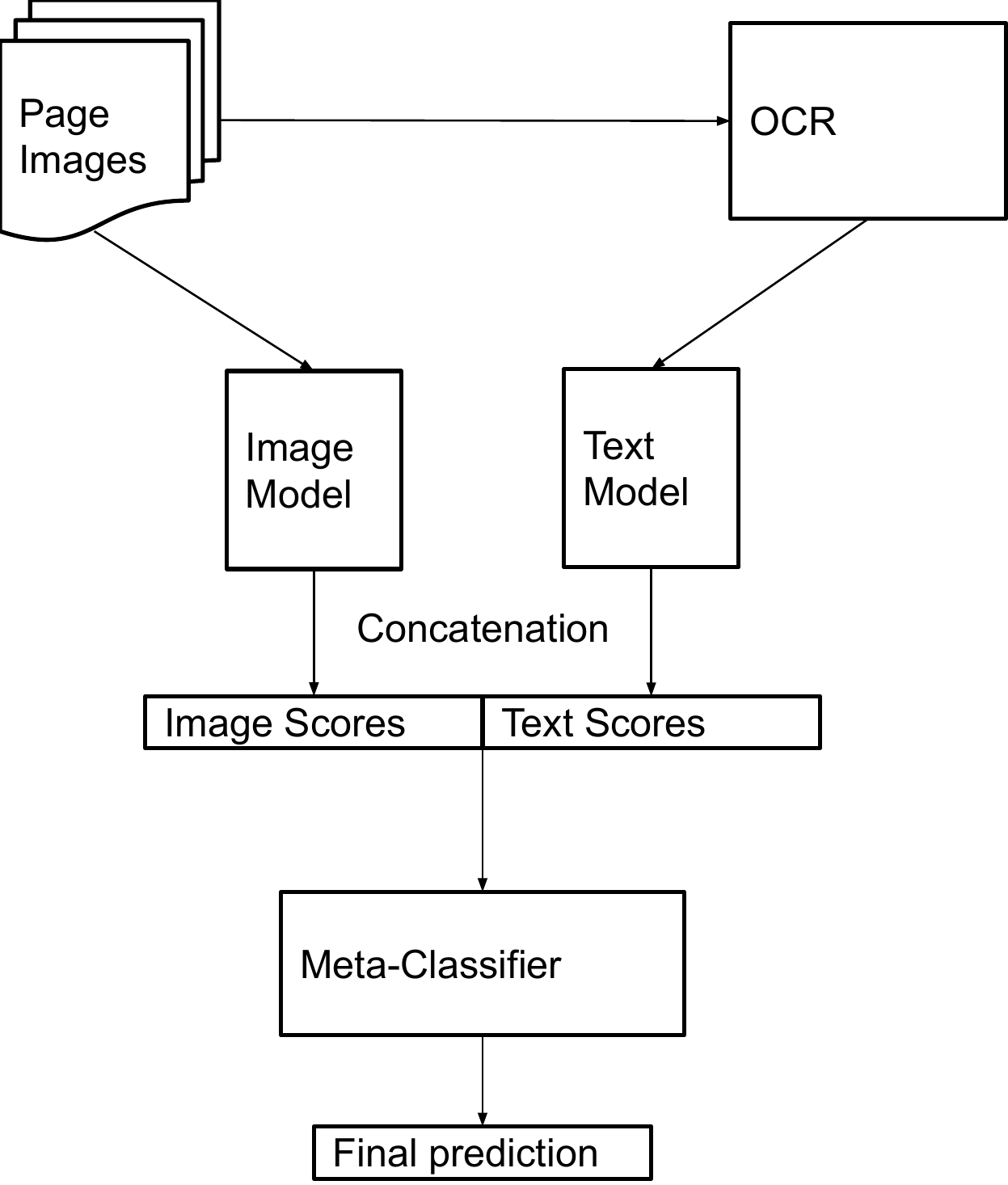}
    \caption{\textbf{Abstract Model Architecture:} All page images are processed with an OCR engine, extracting the text from the image. An Image model is trained on the images themselves, while a text model is trained on this extracted text. The two ``class score" vectors are concatenated and a third meta-classifier is trained on the resulting data. The final class score vector represents the final prediction. }
    \label{fig:abstract_arch}
\end{figure}

We define the abstract structure of the model as having three components, an image classifier, a text classifier, and a meta-classifier that joins the two prior components' predictions into one (Fig. \ref{fig:abstract_arch}). We opt for the ``late fusion" scheme for joining predictions, assuming each classifier has an output of dimension $c$, where $c$ is the number of classes, then our meta-classifier is a mapping from $\R^{2c} \to \R^{c}$. That is to say that the meta-classifier takes two outputs and maps it to one.

The modular nature of this structure allows for the swapping of different classifiers with relative ease and seems to point to a possible generalized procedure for developing page classification modules within a document analysis pipeline. Extending this idea it is easy to imagine that if a new representation was developed (graph representation for example) then one could add a new model trivially without the need of retraining the other two components and only needing to update the meta-classifier.

\subsection{Image Model Architectures}
\begin{figure*}
 \subfloat[Alexnet Architecture\label{fig:alexnet}]{%
  \includegraphics[width=0.35\linewidth]{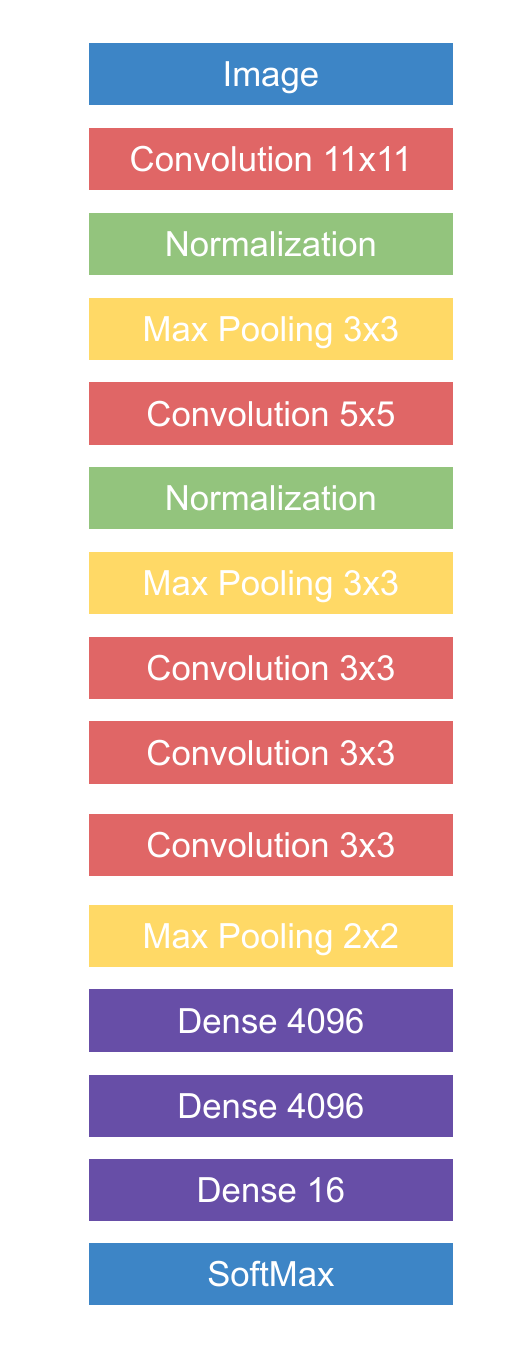}
 }
 \hfill
 \subfloat[VGG16 Architecture\label{fig:vgg16}]{%
  \includegraphics[width=0.45\linewidth]{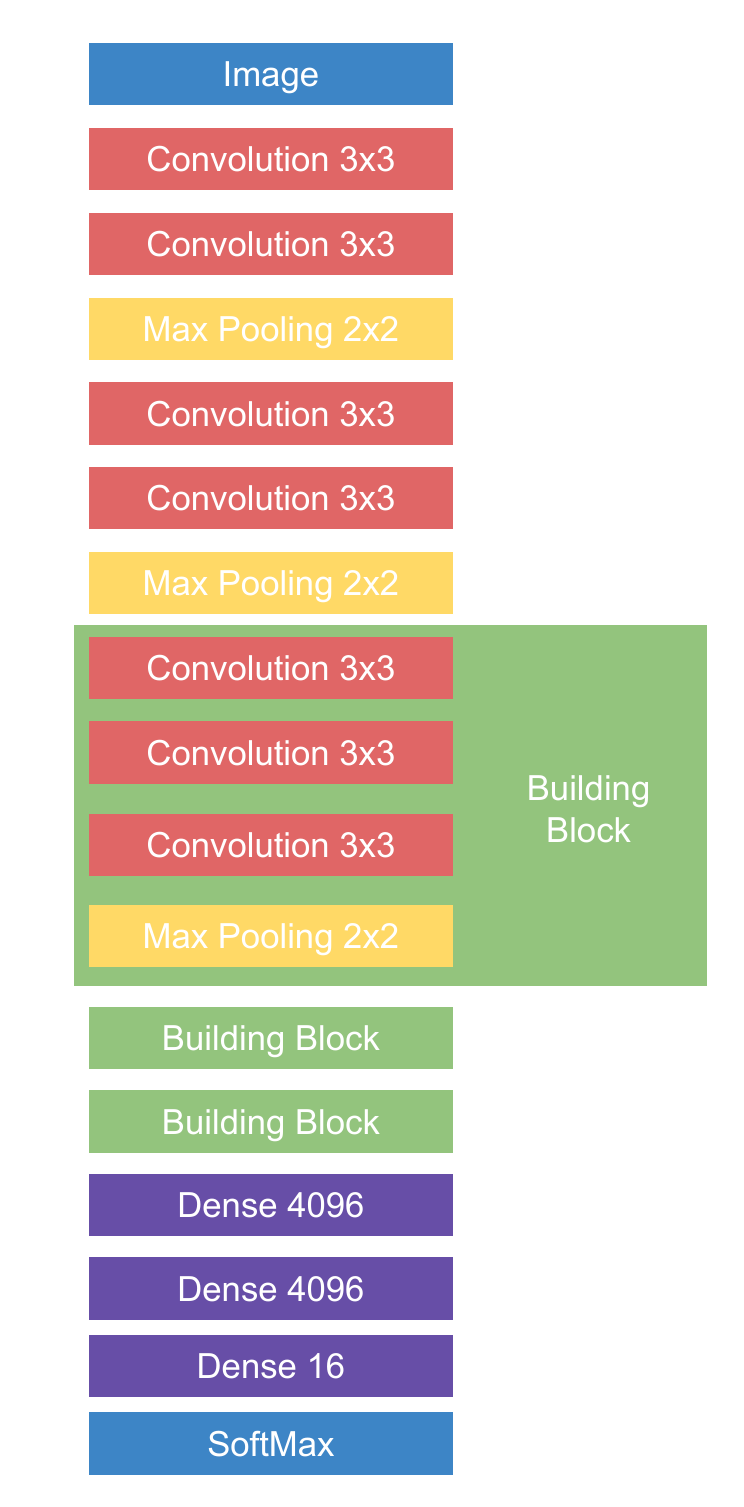}
 }
 \caption{\textbf{Image Model Architectures}}
 \label{fig:img_models}
\end{figure*}

We utilize two standard CNN architectures for the image models, the first is AlexNet (Fig. \ref{fig:alexnet}) with added batch normalization, the second is VGG16 (Fig. \ref{fig:vgg16}). Both models have input dimensions of $227\times227$ and a 16 neuron softmax output layer (corresponding to the 16 classes of RVL-CDIP, similar to those in Afzal et al \cite{Afzal_2017}). Since the source images are grayscale we convert these to RGB and rescale the pixel values to lie within the range $[-1,1]$.

\subsection{Text Model Architectures}
\begin{figure}
    \centering
    \includegraphics[width=7.5cm]{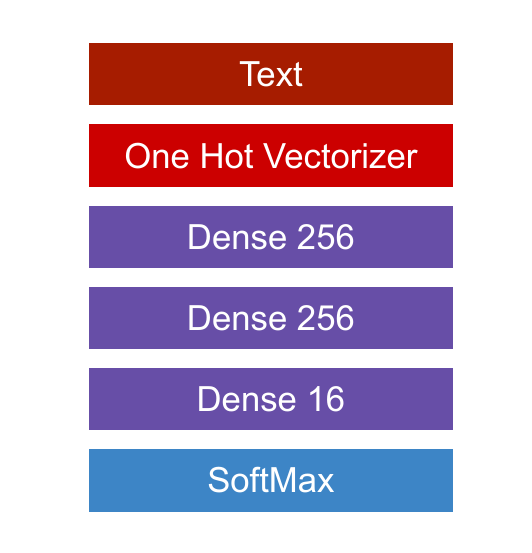}
    \centering
    \caption{\textbf{Text Model Architecture:} Raw text is first one-hot vectorized that is each word in the training set is assigned an index, and every given document is represented as a vector with binary entries corresponding to the presence (or lack thereof) of the corresponding word. These one-hot vectors are then used to train a small dense network.}
    \label{fig:text_model}
\end{figure}

The raw text is first preprocessed into one-hot vectors, that is to say that each document is represented by a binary vector whose components indicate the presence of the word corresponding to that index. These document vectors are fed into a relatively shallow network (see FIG.\ref{fig:text_model}). We denote these as Bag-of-Words (BoW) followed by the number of vocabulary items retained. For example BoW-100K refers to the bag of words model with 100 000 vocabulary words used as features, meaning the input vectors are 100 000 dimensional.

\subsection{Meta-classifier}

The meta-classifier in all experiments is an XGBoost model \cite{Chen:2016:XST:2939672.2939785}. We do not use any regularization parameters instead opting to limit the depth of the trees to control for overfitting (to a maximum depth of 3). The minimal tuning required for this classifier makes it an ideal candidate for meta-classification.

\section{Experiments and Results}

\subsection{Implementation}
All network models are generated using Keras \cite{chollet2015keras} with Tensorflow backend \cite{tensorflow2015-whitepaper}. We also utilize a number of modules from scikit-learn \cite{scikit-learn} to preprocess the text. We take advantage of the XGBoost library for the meta-classifier. We consistently surpass the current state of the art however exact replication with Tensorflow on GPU is a continuing challenge, with many possible sources of non-deterministic behavior\footnote{https://github.com/NVIDIA/tensorflow-determinism}. 

\subsection{Augmentation}
In their study, Tensmeyer et al \cite{tensmeyer2017analysis} suggest that slight shear augmentations ($\theta \in [-10^{\circ} ,10^{\circ} ]$) during training provide the best generalization performance.  We combine these shear augmentations with slight rotations ($\theta \in [-5^{\circ} ,5^{\circ} ]$) in training both the VGG16 and AlexNet models. 
We also note that although 90 degree rotations do not improve performance on this task, in many real-world applications however, this is absolutely necessary as the orientation of the page is not as tightly controlled. Additionally, we experimented with the addition of salt and pepper noise (random minimizing and maximizing of pixels) to simulate scanner effects, this too did not prove to be fruitful in terms of performance. 

\subsection{Optimization}
We utilize SGD with warm restarts \cite{DBLP:journals/corr/LoshchilovH16a}, however we adjust the learning rate over batches as opposed to epochs, essentially reducing to a discontinuous one-cycle learning rate cosine annealing \cite{DBLP:journals/corr/abs-1803-09820} for optimization. The exact decay function is given by:

\begin{equation}
    l(k) = \frac{1}{2}(l_{max}-l_{min})(\cos\left(\frac{k\pi }{N} \right) + 1) + l_{min}
\end{equation}

\begin{conditions}
 l_{max}     &  Initial learning rate. \\
 l_{min}     &  Desired minimum learning rate. \\   
 k &  batch number within the epoch. \\
 N & number of batches per epoch.
\end{conditions}

This policy works well across applications and remains consistent for all models (both image and text) with some adjustment to the bounds (Table. \ref{tab:lr_bounds}). The policy tends to find a strong local minimum however it can accelerate past the best general solution. It may be worthwhile attenuating the schedules peak-to-peak range over epoches or scaling the periodicity as training progresses, however this can decrease the optimizers ability to ``pop out" of local minima.

\begin{table}[h]
    \centering
    \begin{tabular}{|c|c|c|}
        \hline
         Model Type & $l_{max}$ & $l_{min}$  \\
         \hline
         Image & 0.002 & $10^{-6}$ \\
         \hline
         Text & 0.01 & $10^{-6}$ \\
         \hline
    \end{tabular}
    \caption{Learning rate bounds used for the learning rate schedule for training the two types of models.}
    \label{tab:lr_bounds}
\end{table}

\subsection{Results}

As each classifier is trained independently from one another we can see the results of each experiment and the remarkable boost that comes from the combination of different classifiers.

\begin{table}[h]
    \centering
    \begin{tabular}{|c|c|c|}
        \hline
        Model & Validation Accuracy & Test Accuracy \\
        \hline
          AlexNet random init. &  86.29\% &   86.24\% \\
            VGG16 ImageNet init. & \textbf{90.45\%} &   \textbf{90.24\%} \\
        \hline
           BoW-1K &  75.99\% &   76.02\% \\
          BoW-10K &  80.25\% &   80.22\% \\
          BoW-20K &  80.99\% &    80.9\% \\
          BoW-50K &  82.06\% &   81.69\% \\
         BoW-100K &  82.23\% &    82.0\% \\
         BoW-150K &  81.95\% &   81.62\% \\
         BoW-200K &  \textbf{82.42\%} &   \textbf{82.23\%} \\
         BoW-300K &  82.41\% &   82.12\% \\
        \hline
    \end{tabular}
    \caption{Experimental results for component models.}
    \label{tab:component_results}
\end{table}

\begin{table}[h]
    \centering
    \begin{tabular}{|c|c|c|c|}
        \hline
        Image Model & Text Model & Validation Accuracy & Test Accuracy\\
        \hline
        AlexNet &     BoW-1K &  90.79\% &   90.77\% \\
        AlexNet &    BoW-10K &  91.45\% &   91.34\% \\
        AlexNet &    BoW-20K &   91.4\% &   91.32\% \\
        AlexNet &    BoW-50K &  91.41\% &   91.24\% \\
        AlexNet &   BoW-100K &  91.38\% &   91.19\% \\
        AlexNet &   BoW-150K &  \textbf{91.52\%} &   \textbf{91.44\%} \\
        AlexNet &   BoW-200K &  91.42\% &   91.16\% \\
        AlexNet &   BoW-300K &   91.2\% &   91.18\% \\
        \hline
          VGG16 &     BoW-1K &  92.39\% &   92.38\% \\
          VGG16 &    BoW-10K &  92.79\% &   92.67\% \\
          VGG16 &    BoW-20K &  92.97\% &   92.84\% \\
          VGG16 &    BoW-50K &  \textbf{93.08\%} &   92.94\% \\
          VGG16 &   BoW-100K &  93.06\% &   92.92\% \\
          VGG16 &   BoW-150K &  92.86\% &   92.81\% \\
          VGG16 &   BoW-200K &  93.07\% &   \textbf{93.03\%} \\
          VGG16 &   BoW-300K &  93.05\% &   \textbf{93.03\%} \\
        \hline
    \end{tabular}
    \caption{Experimental results for multimodal models.}
    \label{tab:multimodal_results}
\end{table}

\begin{table}[h]
    \begin{tabular}{|l|p{1.5cm}|p{3cm}|} \hline
        Source & Reported Test Accuracy & Comments \\ \hline
        Afzal et al \cite{Afzal_2017} & 90.97\% & Single well tuned VGG16 intialized on pretrained ImageNet weights. \\ \hline
        Das et al \cite{das2018document} & 91.11\% & Single well tuned VGG16 intialized on pretrained ImageNet weights. \\ \hline
        Das et al\cite{das2018document} & 92.21\% & Ensemble of holistic and region based VGG16s.\\ \hline
        Proposed Work & 93.03\% & Ensemble of VGG16 and MLP based BoW models. \\ \hline
        Proposed Work & 93.07\% & Ensemble of all component models. \\ \hline
    \end{tabular}
    \caption{Comparison of past reported test accuracies with proposed work.}
    \label{tab:past_results}
\end{table}

We see that even the addition of a low ``resolution" bag-of-words model can generate significant lift to the image models superior performance. It is also interesting to note that the combination of randomly initialized AlexNet and BoW-10K beats out the best reported test accuracy for a single image classifier \cite{das2018document}, exceeding the performance of the well tuned VGG16. While the best performing model consists of a VGG16 image component and 200 000 word text model, with a test accuracy of 93.03\%.

The modular nature of this architecture also allows for the simultaneous ensembling of all the component models, resulting in a validation and test accuracy of 93.12\% and 93.07\% respectively. Although an interesting result, this type of ensembling is likely not practical in an industrial scenario due to the requirement of evaluating the 10 component model and single ensemble model.

\section{Conclusion}
It is clear from the results that the inclusion of extracted text in the development of document classification models improves the quality and accuracy of predictions. The proposed method exceeds the current state-of-the-art for test accuracy on the RVL-CDIP dataset and sets a new standard for document classification methods to be compared to. 

The work here only takes advantage of a bag-of-words approach to the text classification component, a further avenue for research could include extending the more recent embedding approaches to account for transcription errors.

\section{Appendix}

\subsection{RVL-CDIP Data Quality}
\begin{figure}[h]
    \centering
    \frame{\includegraphics[width=0.4\linewidth]{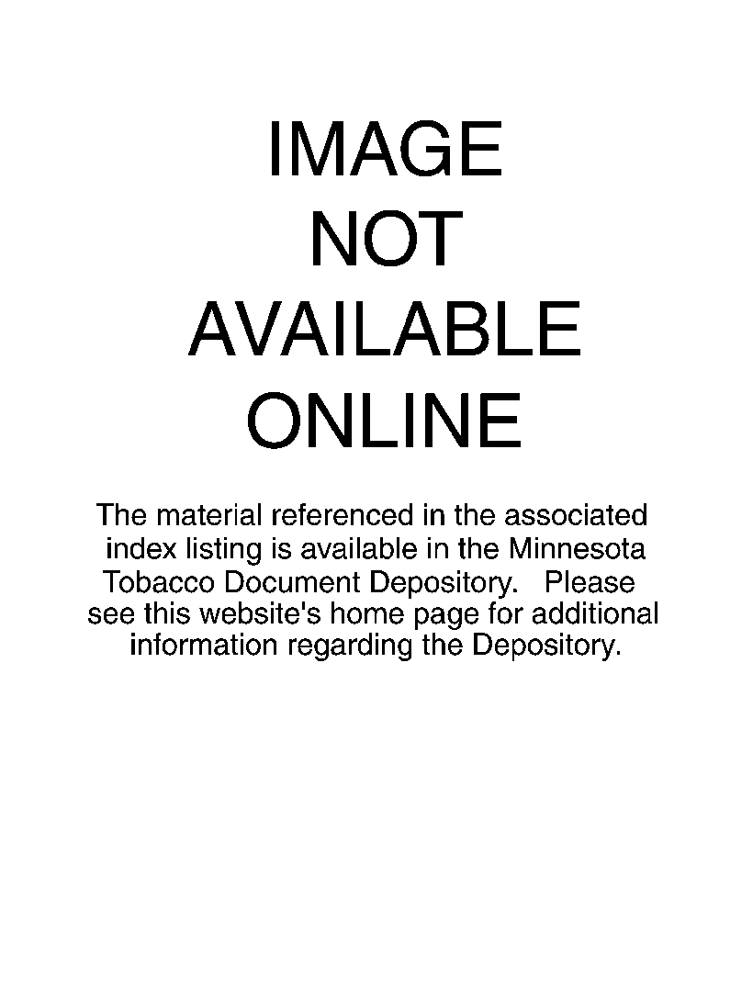}}
    \caption{This image occurs at least 426 times in the training (373 instances) and testing (53 instances) sets. It is spread across classes according to the accompanying table (Table. \ref{tab:duplicate_classes}). As the text suggests it is likely due to a fetching error in the original datasets creation.}
    \label{fig:image_not_found}
\end{figure}

The open RVL-CDIP dataset suffers from some data quality issues, namely duplicated images across sets (training, testing, and validation) and classes. i.e. the same image can occur across classes and sets. The most obvious example of this type of image is illustrated in figure \ref{fig:image_not_found}. Although further study is requred into the data quality of RVL-CDIP, the problem does not seem to be far reaching with an estimated upper bound of 2259 duplicate images. We arrived at this number by examining the unique texts extracted from Tesseract. A more thorough examination is required (potentially with an image hashing technique) to establish the true number of duplicated images.

\begin{table}
    \begin{tabular}{c|c}
        Class & Count of Duplicate Image \\
        \toprule
        4  &    322 \\
        9  &     30 \\
        12 &     22 \\
        5  &     22 \\
        1  &     21 \\
        10 &      3 \\
        15 &      2 \\
        13 &      1 \\
        11 &      1 \\
        7  &      1 \\
        0  &      1 \\ \hline
        Total & 426
    \end{tabular}
    \caption{Class breakdown of ``image not available " duplicate images.}
    \label{tab:duplicate_classes}
\end{table}

\newpage

\bibliographystyle{ieeetr}
\bibliography{refs}

\end{document}